\documentclass[journal]{IEEEtran}
\usepackage{float}

\ifCLASSINFOpdf
\else
\fi
\usepackage{mathrsfs}
\usepackage{amsmath}
\usepackage{multirow}
\usepackage{graphicx}
\usepackage{color}
\usepackage{caption}
\usepackage{tabularx}
\usepackage{booktabs}
\usepackage{stfloats}
\usepackage{subfloat}
\usepackage{amsthm,amsmath,amssymb}
\usepackage{subfig}
\usepackage{amssymb}
\usepackage[citecolor=blue]{hyperref}

\begin{document}
    %
    \title{Dynamic Message Propagation Network for RGB-D Salient Object Detection}
    %
    %
    %
    
    \author{Baian~Chen,
            Zhilei~Chen,
            Xiaowei~Hu,
            Jun~Xu,
            Haoran~Xie,
            Mingqiang~Wei,
            Jing~Qin}
    
    %
    %

    \markboth{Journal of \LaTeX\ Class Files,~Vol.~14, No.~8, August~2021}%
    {Shell \MakeLowercase{\textit{et al.}}: Bare Demo of IEEEtran.cls for IEEE Journals}
    %



    \maketitle
    
    \begin{abstract}
    This paper presents a novel deep neural network
    framework for RGB-D salient object detection by controlling the message passing between the RGB images and depth maps on the feature level and exploring the long-range semantic contexts and geometric information on both RGB and depth features to infer salient objects.
    To achieve this, we formulate a dynamic message propagation (DMP) module with the graph neural networks and deformable convolutions to dynamically learn the context information and to automatically predict filter weights and affinity matrices for message propagation control.
    We further embed this module into a Siamese-based network to process the RGB image and depth map respectively, and design a multi-level feature fusion (MFF) module to explore the cross-level information between the refined RGB  and depth features.
    Compared with 17 state-of-the-art methods on six benchmark datasets for RGB-D salient object detection, experimental results show that our method outperforms all the others, both quantitatively and visually.

    \end{abstract}
    

    %
    \IEEEpeerreviewmaketitle

    \section{Introduction}
\begin{figure}[t]
        \centering
        \includegraphics[width=0.47\textwidth]{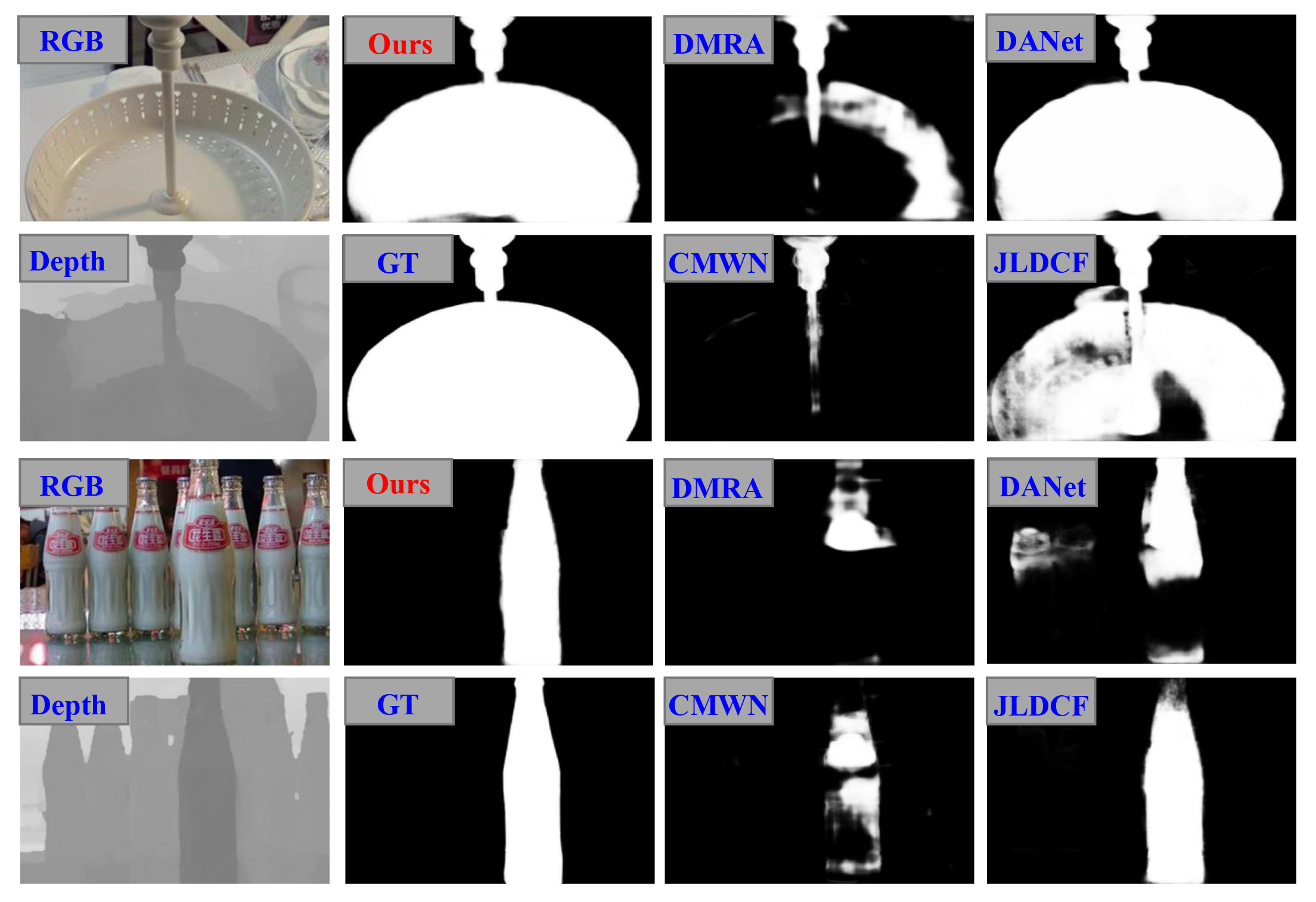}
        \caption{Saliency prediction by RGB-D SOD methods (i.e., DMRA~\cite{9010728}, DANet~\cite{zhao2020single}, CMWNet~\cite{li2020crossmodal}, JLDCF~\cite{9157592} and Ours). Our DMP-Net generates better saliency maps than other methods, since the long-range dependencies from our dynamic message propagation (DMP) module can reserve the boundary information of the salient object to the greatest extent.}
        \label{fig:fig1}
    \end{figure}

    %
    %
    %
    %
    \IEEEPARstart{S}{alient} object detection (SOD) aims at detecting the most attractive objects in a scene, which contains lots of vision applications, such as stereo matching~\cite{8953688}, image understanding~\cite{6888473}, co-saliency detection~\cite{9358006}, video detection and segmentation~\cite{8047320}, medical image segmentation~\cite{9098956}, and person reidentification~\cite{7289409}. 
    RGB-D images provide the additional depth information to improve the performance of SOD in the complex scenes, especially when there exists (1) low contrast between the foreground and the background; and (2) similar objects in the background, as shown in the first column of Fig.~\ref{fig:fig1}.
    
    To explore the depth information, the current methods~\cite{9157592,9010728,zhao2020single} leverage the convolutional neural networks (CNNs) with various multi-modal and multi-level strategies to combine RGB and depth maps for boosting the performance of RGB-D SOD.
    However, these CNN-based algorithms adopt kernels with the fixed perceptive fields to capture the local information, which is hard to obtain the long-range contextual dependencies between the RGB image and depth map, thus limiting the performance of RGB-D SOD.

    To solve the above issues, we present to dynamically control the message passing between the RGB features and depth features, and simultaneously capture the long-range contextual information on both RGB and depth features to infer the salient objects by exploring the long-range contexts and geometric information on RGB images and depth maps, respectively. 
    To achieve this, we formulate a dynamic message propagation (DMP) module based on the graph neural networks (GNNs), which adaptively samples surrounding contexts and depth-aware information from both RGB and depth feature maps and dynamically predicts filter weights and affinity matrices for message propagation control. 
    Thus, it is fully different from the regular kernels in CNNs, which have the limited capability to model long-range dependencies and only perceive the fixed local regions. 
    Besides, we design a multi-level feature fusion (MFF) module to combine the multi-level features from the network decoder
    to make full use of the cross-level information between the refined RGB features and depth features.
    Finally, we evaluate the proposed framework on six widely-used benchmark datasets for RGB-D salient object detection and the experimental results show that our method achieves state-of-the-art performance on all the datasets in terms of almost all the evaluation metrics.
    
    Our contributions are be summarized as follows:
    \begin{itemize}
        \item We design a dynamic message propagation network (DMP-Net) for RGB-D salient object detection by formulating a DMP module to dynamically sample context- and depth-aware features 
        and further explore the long-range contextual dependencies between the RGB images and depth maps.
        \item We design a multi-level feature fusion (MFF) module to leverage the cross-modal characteristics between the RGB images and depth maps and to enhance the comprehensive semantics for feature reconstruction.
        \item We conduct a comprehensive comparisons with seventeen state-of-the-art RGB-D SOD methods on six benchmarks datasets.  Experimental results show that our DMP-Net achieves the best performance on all the benchmark datasets and shows a great improvement on the challenging cases.
    \end{itemize}
    
    The remainder of this paper is organized as follows. In Section \ref{sec:rel}, we briefly introduce the related works on RGB-D saliency object detection (SOD) and graph neural networks (GNNs). In Section \ref{sec:met}, we present the overview framework and the details of the proposed DMP-Net. In Section \ref{sec:exp}, we show the experimental results on six datasets and compare our DMP-Net with seventeen state-of-the-art RGB-D SOD methods. Lastly, we present the conclusion in Section \ref{sec:con}.

    \section{Related Work}
    \label{sec:rel}
    In this section, we briefly review the related progress of RGB-D salient object detection and graph neural network in recent years.
    
    \subsection{RGB-D Salient Object Detection}
    Depth maps have several significant attributes like boundary cues, spatial structure and layout information, which are crucial for boosting the performance of SOD in complicated scenes. In the past years, traditional RGB-D SOD methods mainly develop hand-crafted features~\cite{7301391,7025222,7780626,7938352}. These hand-crafted features usually focus on contrast-based cues like color, edge, depth and region.
    However, traditional methods cannot usually achieve satisfactory performance of SOD, due to the lack of well-designed fusion frameworks like CNNs to integrate RGB and depth features in a multi-level manner. 
    
    With the development of deep neural networks (DNNs), this problem has been alleviated. Many methods have attempted to apply DNNs to their models to improve the performance of SOD and achieved encouraging results~\cite{7879320,8953818,9010728,8594373}. 
    
    According to the different ways of feature fusion, existing RGB-D SOD methods can be roughly divided into three groups.\\
    (1) Early-stage and late-stage fusion. They are both simple and intuitive. The initial early-stage fusion methods directly concatenate RGB images and depth maps as a four-channel input and handle them in an equal manner~\cite{7301391,7938352,Liu_2019}. One of the improved strategies is feeding RGB and depth images into a plain network separately to achieve low-level representations from multi-modality, and then feeding these combined cross-modal representations into a subsequent network for further processing~\cite{7879320}.
    Contrary to early fusion, late-stage fusion utilizes two parallel network to process RGB and depth streams separately. The methods of fusing the high-level features from different convolutional neural networks can be divided into two categories. One is directly concatenating them to generate the final saliency prediction~\cite{8091125,8704255}. The other is firstly obtaining individual saliency maps from RGB and depth streams, and then combining them for a further final saliency map~\cite{Yu_2019}.\\
    (2) Middle-stage fusion. Middle-stage fusion is the most effective and most widely-used strategy for RGB-D SOD, since it can completely explore the complementary between RGB images and depth maps~\cite{9247470,8603756,pang2020hierarchical,zhai2021bifurcated}. This strategy mainly extracts features from the RGB stream and the depth stream in different layers of the encoder network, then fuses them in a subsequent decoder network to generate the final saliency map. ICNet~\cite{9024241} designs an information conversion module to fuse high-level cross-modal features in an adaptive manner. Furthermore, a cross-modal depth-weighted combination (CDC) block is proposed to enhance RGB features with depth cues in different layers. JL-DCF~\cite{9157592} attempts to jointly train RGB images and depth maps in a Siamese network and proposes a densely-cooperative integration strategy to combine the multi-modality and multi-level features in the decoding stage. DPANet~\cite{9247470} designs a gated multi-modality attention (GMA) module to obtain the long-range dependencies with a gated controller. In addition, this model can control the fusion rate by introducing a gated function to avoid disturbances from the uncertain depth maps. \\
    (3) Multi-scale fusion. Multi-scale fusion is the further enhancement based on middle-stage fusion, which takes multi-scale RGB and depth features as input in a convolutional module to capture cross-scale complementarities and learn context-aware feature representations \cite{9321705,li2020crossmodal,9366409}. BiANet~\cite{9321705} proposes a multi-scale bilateral attention module (MBAM) to exploit global salient information in a multi-level manner. DSA$^2$F \cite{sun2021deep} proposes a depth-sensitive RGB feature modeling scheme using the depth-wise geometric prior of salient objects to enhance RGB features and an automatic multi-modal multi-scale fusion module to fuse the enhanced RGB features and the depth features. Our method takes multi-scale depth features as the input of the DMP module to make full use of the geometric information of different perceptive field.
    
    \begin{figure*}[t]
        \centering
        \includegraphics[width=\linewidth]{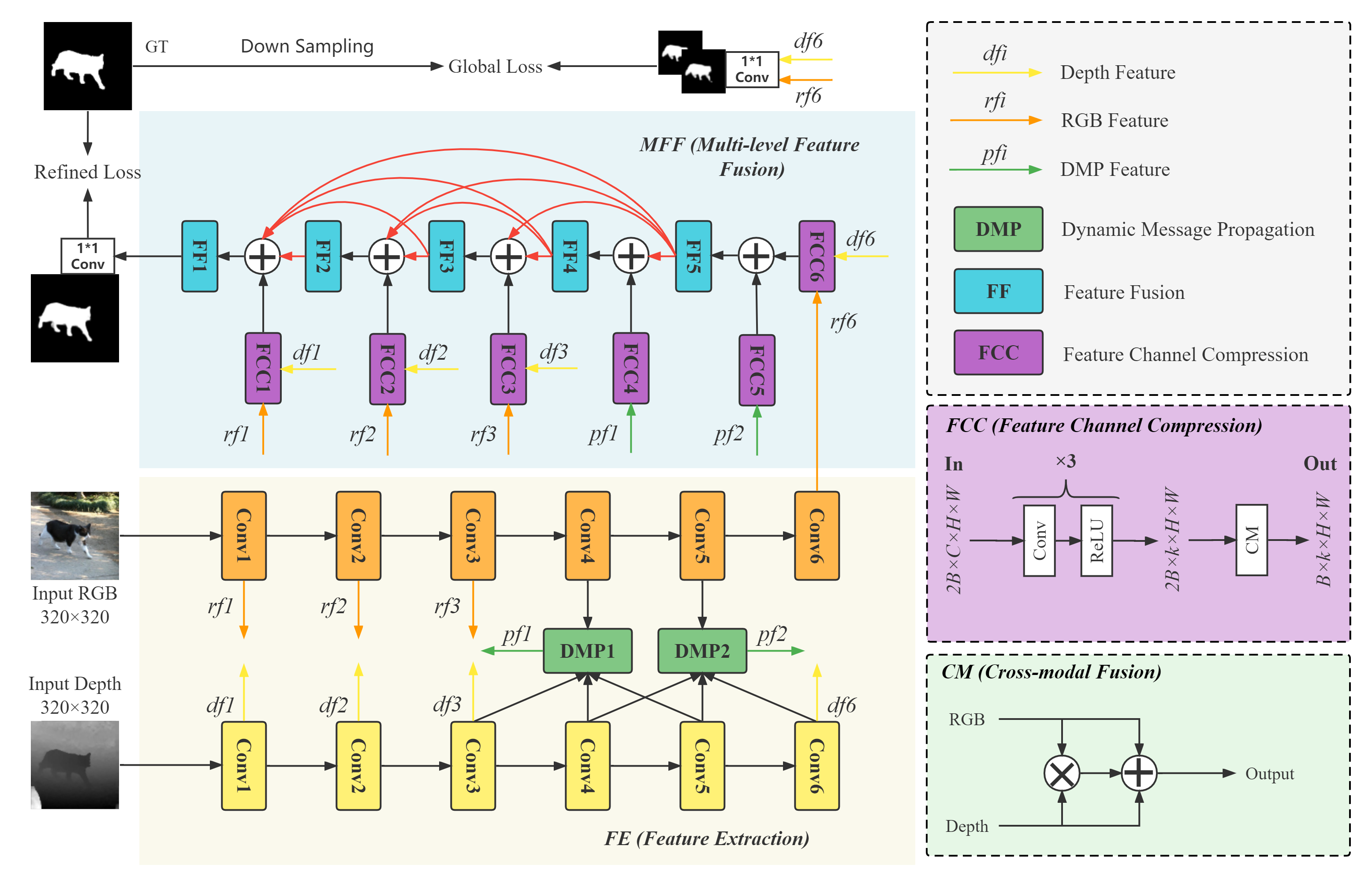}
        \caption{\textbf{The Pipeline of our proposed DMP-Net.} The encoder (the left part) consists of a simple VGG layer and the complete DenseNet (the rest layers). We introduce the proposed dynamic message propagation (DMP) modules in the high-level layers. The decoder (the right part) mainly consists of multi-level feature fusion (MFF) modules for multi-modal and multi-level feature fusion.}
        \label{fig:fig2}
    \end{figure*}
    
    \subsection{Graph Neural Network.} 
    The graph neural networks (GNNs) \cite{4700287} is a novel architecture and discriminated from the convolutional neural networks. It is composed of nodes and edges, which can effectively model long-range contextual dependencies by propagating information along the graph-structured input data. Benefiting from this characteristic, graph neural networks are widely used in many tasks ~\cite{kipf2017semisupervised,9378450,9324530,9177430,wang2021depthconditioned}. 
    
    The message passing algorithm was first proposed in the clustering modal, the essence of which is propagating the message of each element to neighbours, and after several iterations of information exchange, a stable cluster map is finally generated.
    Message passing neural networks (MPNNs) \cite{gilmer2017neural} proposes a GNNs based framework by learning a message passing algorithm and aggregation procedure to compute a function of their entire input graph for quantum chemistry. Subsequently dynamic graph message passing networks (DGMP-Net) \cite{9157180} attempt to regard images as a graph to apply MPNNs on the image processing task for modelling long-range dependencies.
    
    In this paper, we propose a graph-structured module that utilizes deformable convolutional networks to learn the contextual feature representations from RGB and depth streams and propagates these messages between RGB and depth maps for RGB-D SOD.
    
    \section{Methodology}
    \label{sec:met}
    In this section, we first introduce the pipeline of DMP-Net, and then describe the dynamic message propagation (DMP) module and the multi-level feature fusion (MFF) in detail, followed by the implementation of DMP-Net.
    
    \subsection{Overview}
    \textbf{Motivation}. 
    It is crucial to model the long-range contextual dependencies for the scene understanding tasks such as semantic segmentation and object detection. As a result of the fixed convolutional kernels of CNNs, they are limited in acquiring the long-range dependent relationships. Inspired by \cite{9157180}, we can adopt the graph neural network to learn a dynamic feature representation of salient objects, which depends on flexible nodes and edges to replace the fixed kernels to learn the long-range contextual information. Moreover, multi-scale depth features including rich geometric information can assist with RGB features in understanding the object shape in a coarse-to-fine manner.
    
    Based on the motivation, we present the framework of our DMP-Net in Fig.~\ref{fig:fig2}. The backbone of DMP-Net is a typical two-stream encoder-decoder architecture, where RGB and depth features are processed separately and then fused in a cross-modality and cross-level manner. 
    We adopt the Siamese network as the encoder network. It possesses the characteristic of sharing parameters, which facilitates to exploit the commonality between the RGB and depth modalities. 
    In the encoder, in order to model the long-range contextual dependencies, we apply two dynamic message propagation (DMP) modules on the high-level features to aggregate the RGB features and multi-scale depth features. In the decoder, features in each layer are fused with the features of the corresponding layer in the encoder network via skip connections to revive the original scale information. What is more, each layer is connected with all higher-level layers to capture rich multi-scale information. 
    
    \subsection{Feature Extraction}
    With the tremendous progress of convolutional neural networks, a plenty of feature extraction backbone networks \cite{he2016deep,simonyan2014very,huang2017densely} are proposed, which have been universally applied in many models for different tasks. For the reason that the RGB image and depth map contain different channels, we first normalize the depth map and repeat it in channel dimension. To extract rich informative features, DenseNet \cite{huang2017densely} is adopted as the backbone network, in which we retain all convolutional layers and discard the last pooling layer and fully connected layer. However, it is unsatisfactory that the first convolutional layer in DenseNet decreases the resolution of image, leading to incomplete features. Similar to \cite{9157592}, we borrow the first convolution layer from VGG-16 \cite{he2016deep} to extract features of original resolution. It is worth noting that we leverage a Siamese strategy to excavate RGB and depth features simultaneously.
    
    \begin{figure}[t]
        \centering
        \subfloat[Dynamic message propagation (DMP) module]{
            \includegraphics[width=0.8\linewidth]{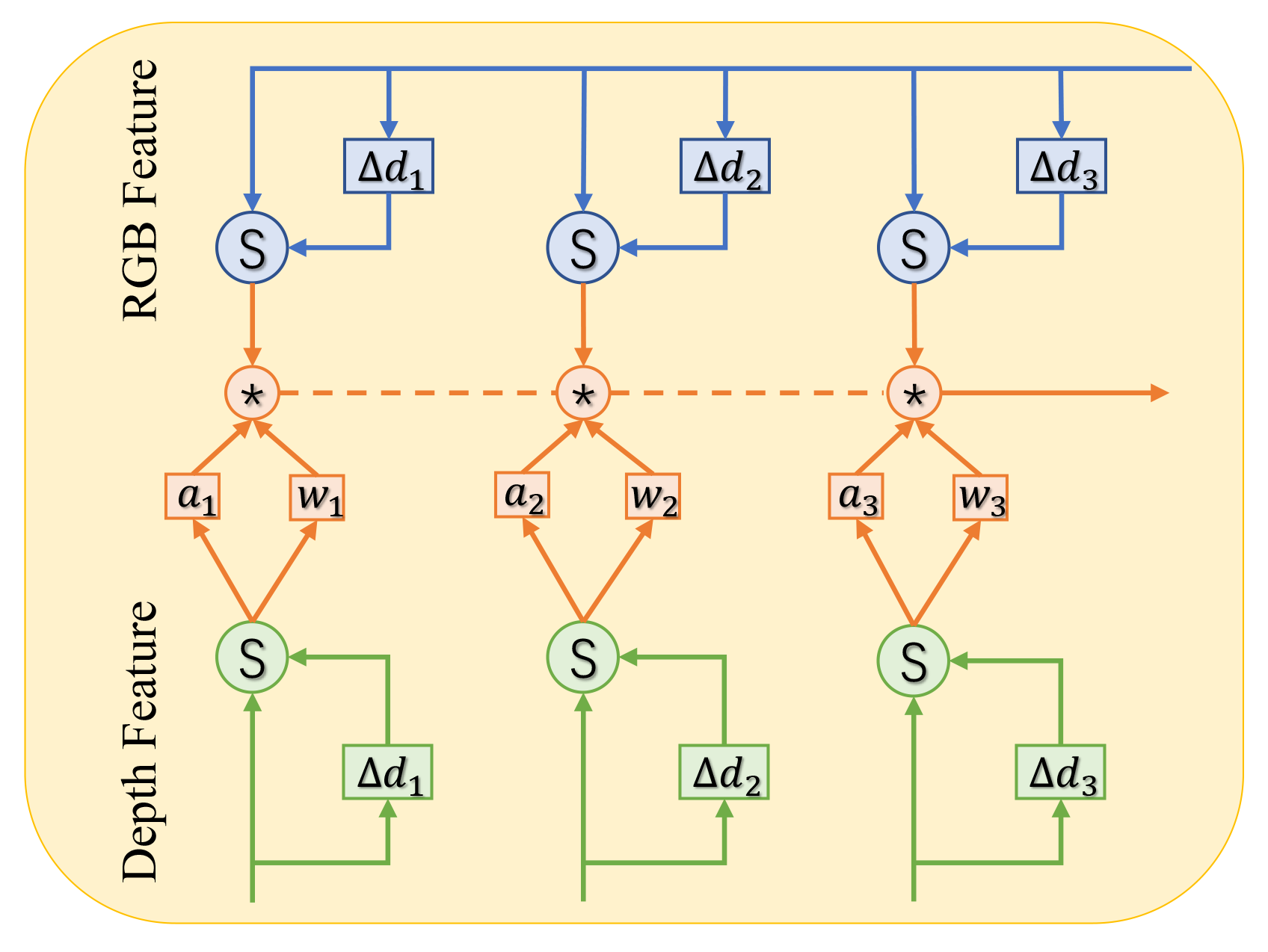}
        }
        
        \subfloat[One branch in DMP]{
            \includegraphics[width=0.99\linewidth]{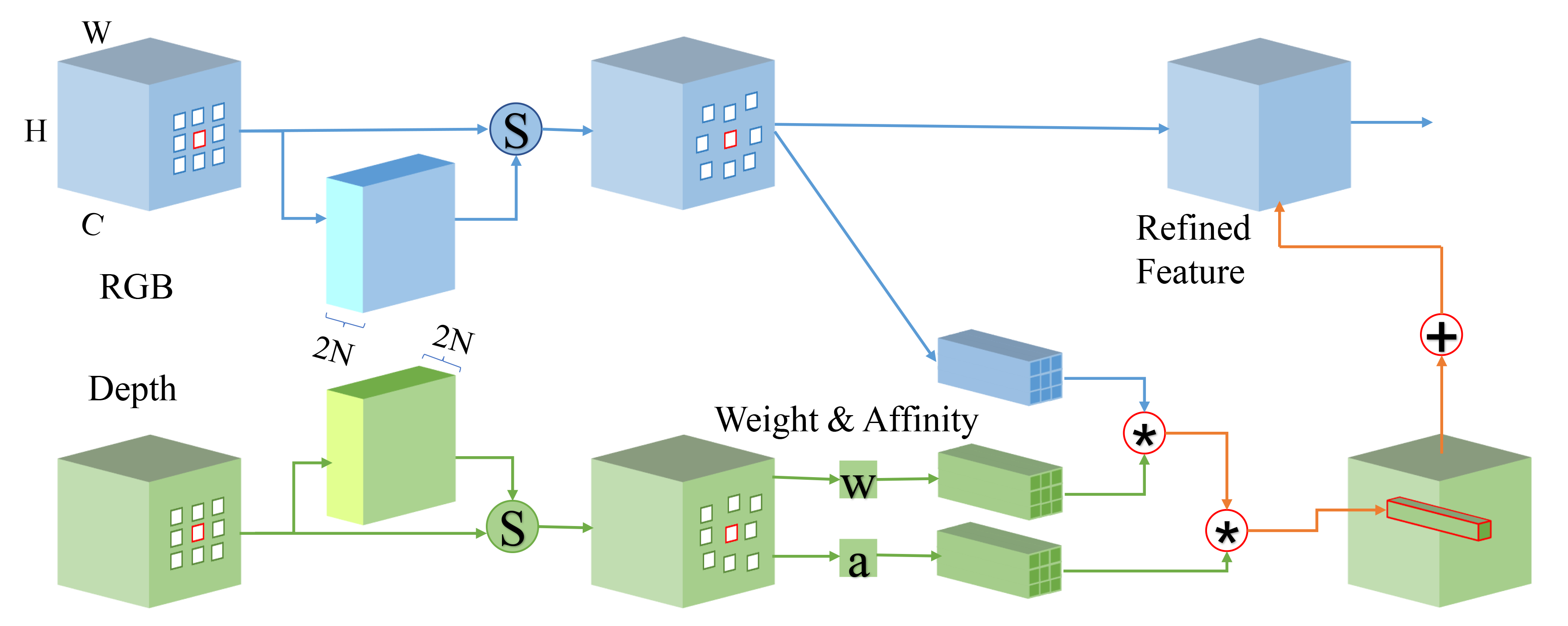}
        }
        \caption{DMP consists of three branches (a) for dynamically fusing the RGB features and multi-level depth features. Each branch (b) will sample dynamic nodes from the image and depth feature graphs, and learn the filter weights and affinity matrices from depth nodes to propagate the message. $\Delta d_i$, $a_i$ and $w_i$ represent the learned walk, filter weights and affinity matrices for each layer separately. $S$ represents sampling context- and depth-aware nodes.}
        \label{fig:fig3}
    \end{figure}
    
    \subsection{Dynamic Message Propagation (DMP)}
    The message passing mechanism constructs a feature graph $G = \{V, E, A\}$, where \textit{V} as the node set, \textit{E} as the edge set, and \textit{A} as an adjacency matrix. 
    DMP regards the input feature maps as a graph, in which each pixel is a vector node $v_i\in{R^C}$, and all nodes of the graph make up the set $V=\{v_i\}^N_i$, where $C$ is the channel of the input feature maps, $N$ is the total number of the pixels. 
    The binary or learnable adjacency matrix $A \in {R^{N\times{N}}}$ can describe the connections between nodes in a self-loops manner. 
    
    The goal of this mechanism is to refine the latent feature vectors $h_i$ by excavating hidden structured information from the feature vectors at different node locations. The message passing phase generally takes $T$ iteration steps to refine the latent feature vectors and consists of the message calculation step $M^t$ and message updating step $U^t$. Considering a latent feature vector $h^{(t)}_i$ at iteration $t$, it dynamically samples K nodes which are connected to form a local field $v_i \subset{V}$ and $v_i\in{R^{K\times{C}}}$, where $C$ is the dimension of the vector and $K\ll{N}$. Thus, the message calculation step for the node $i$ is formulated as:
    \begin{equation}
        \begin{aligned}
            m^{t+1}_i &= M^t(A_{i,j}\{h^{(t)}_1,...,h^{(t)}_K\}, w_j)\\
            &= \sum_{j\in{N(i)}}A_{i,j}h^{(t)}_jw_j
        \end{aligned}
        \label{eq1}
    \end{equation}
    where $A_{i,j}$ describes the connection relationship between the latent nodes $h^{(t)}_i$ and $h^{(t)}_j$, $N(i)$ denotes all of $K$ sampled nodes from $v_i$, and $w_j\in{R^{C\times{C}}}$ is a transformation matrix to calculate message on the hidden node $h^{(t)}_j$. Then, the node $h^{(t)}_i$ will be updated in the message updating step $U^t$ with a residual addition of calculated message and the preceding node status:
    \begin{equation}
        h^{t+1}_i=U^t(h^{t}_i,m^{t+1}_i)=\sigma(h^{t}_i+\alpha^m_im^{t+1}_i)
        \label{eq2}
    \end{equation}
    where $\alpha^m_i$ is a learnable parameter to scale the message, and the operation $\sigma(\cdot)$ is a non-linearity function such as ReLU. After $T$ iterations, the final refined features will be obtained by message propagation on each node.
    
    DMP consists of two stages. The first is node sampling on RGB-image and depth-map features for selecting the most object-relevant nodes in the graph. The second is to generate hybrid filter weights and affinity matrices from depth sampled nodes to enrich the contextual information of the image features. 
    
    As shown in Fig.~\ref{fig:fig3}, we first sample context-aware nodes in both image and depth feature maps. We try to leverage the fixed sampling strategy and find it difficult to be adapted to the complicated scenes where salient objects have different scales. Thereby, we adopt deformable convolution \cite{8237351,8953797} to dynamically sample nodes. 
    For each node $v_i$, the sampling number $K$ represents the receptive field of $v_i$ similar to the convolutional kernel size. We assign the learned walk $\Delta{d_{i,j}}\in{R^D}$ to $K$ nodes around $v_i$ as the moving path relative to $R$, where $j\in{N(i)}$ denotes $K$ sampled nodes around $v_i$, $D = 2$ is the 2D space offset, and $R$ is the fixed $3\times{3}$ receptive field of $v_i$,
    \begin{equation}
        R = \{(-1,-1),(-1,0),...,(0,1),(1,1)\}
        \label{eq3}
    \end{equation}
    We apply a convolutional layer to produce the offset field with $2K$ channels, where each element represents $K$ predicted walks of the sampled nodes. Then, we add the extra offset to each location in $R$,
    \begin{equation}
        \Delta{d_{i,j}} = p_i + p_j + \Delta{p_j}
        \label{eq4}
    \end{equation}
    where $p_i$ is the location of $v_i$, $p_j$ is the location of sampled node $v_j$ in $R$, $\Delta{p_j}$ is the predicted walk of $v_j$.
    
    After the above operations, we acquire the dynamic sampled nodes on both image and depth feature maps. Due to the difference between RGB and depth modalities, we predict the walk for each feature map independently. We sample three different-level depth nodes for obtaining the comprehensive depth-aware context. Subsequently, we generate the affinity matrix $A$ and the transformation matrix $W$ based on multi-level depth sampled nodes, and calculate with image nodes separately. 
    For the level $l$, we have acquired the sampled nodes $\hat{v}_{i,j}$ and $\tilde{v}_{i,j}$ from the image and depth feature graphs respectively. Different-level depth features have been aligned the corresponding scale with the RGB features via down-sampling or up-sampling operations. The depth nodes $\tilde{v}_{i,j}$ further generate the affinity matrix $A^l_{i,j}$ and the transformation matrix $W^l_{i,j}$ by applying $3 \times 3$ convolutional layers, subsequently calculated with the sampled image features like Eq. \ref{eq1}:
    \begin{equation}
        m^{t+1}_i = \sum_{l\in{L}}{\sum_{j\in{N(i)}}\beta_l{A^l_{i,j}\hat{h}^{l,(t)}_{j}W^l_{i,j}}}
        \label{eq5}
    \end{equation}
    where $L$ denotes the layers that need message propagation, $\hat{h}^{l,(t)}_{j}$ is the feature vector which is the image node $\hat{v}_{i,j}$ from the stage $l$ with the walk $\Delta{d_{i,j}}$, and $\beta_l$ which is set to 1 to balance depth feature maps from different levels.
    
    \subsection{Multi-level Features Fusion (MFF)}
    We first design a Feature Channel Compression (FCC) module (as shown in Fig.~\ref{fig:fig2}) to compress all extracted features to unified channels $k$ for subsequent procession. The predictions of the last level are supervised by the ground truth and we denote the loss function as $ L_g $.
    
    In order to efficiently carry out multi-level feature fusion, we fuse the corresponding RGB features $X^l_{rgb}$ and depth features $X^l_d$ in the same layer to obtain the fused features $X^l_f$. This cross-modal fusion module consists of simple element-wise addition and multiplication, which is formulated as:
    \begin{equation}
        X^l_f = X^l_{rgb} \oplus X^l_d \oplus (X^l_{rgb} \otimes X^l_d)
        \label{eq6}
    \end{equation}
    where $l$ represents the index of the encoder layer, and $\oplus$ and $\otimes$ denote the element-wise addition and multiplication respectively. The element-wise addition $\oplus$ emphasizes the cross-modal complementarity, while element-wise multiplication $\otimes$ focuses more on the cross-modal commonality. Because DMP is leveraged in the forth and fifth layers of the encoder, which dynamically aggregates the RGB features and multi-level depth features, there is no need to fuse features of this two layers again. The fused features are subsequently sent to the feature fusion (FF) module in each layer.
    
    \begin{figure}
        \centering
        \includegraphics[width=\linewidth]{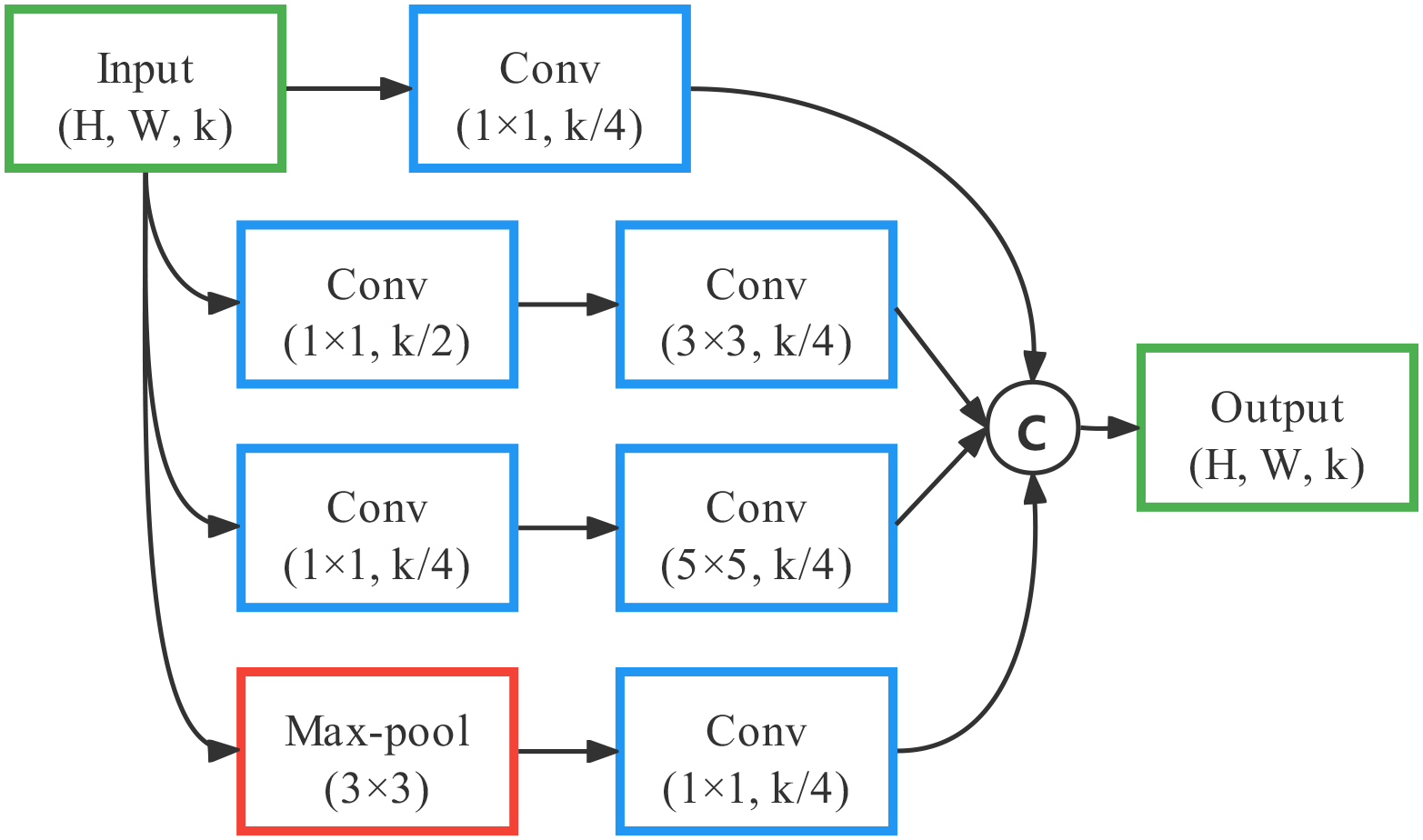}
        \caption{\textbf{The structure of Multi-scale Feature Fusion (FF) module.} All convolutional layers and the max-pooling layer have $k/4$ output channels, which aims to maintain the number of feature channels.}
        \label{fig:fig4}
    \end{figure}
    
    Different from the traditional UNet-like \cite{ronneberger2015unet} decoder, we utilize dense link strategy to aggregate and enrich multi-level features, which introduces all higher-level FF features and the corresponding fused side-out features as the input of FF with simple element-wise addition.
    
    Inspired by the Inception \cite{7298594} structure, FF applies four branches to the input features to extend the receptive field (see Fig.~\ref{fig:fig4}), and concatenate them to obtain the informative features. Specifically, MFF includes multi-scale convolutions with filter sizes of $1 \times 1$, $3 \times 3$, and $5 \times 5$, and the max-pooling operation to capture fine-grained details. Besides, the features of the last MFF are sent to a prediction layer to obtain the final saliency map. We denote the loss function as $ L_r $.
    
    \subsection{Loss Function}
    The total loss function is formulated by the global prediction loss $L_g$ and the refined prediction loss $L_r$. We apply a $1\times1$ convolutional layer on the last encoder layer to obtain the global prediction maps $S^r$ and $S^d$, and another $1\times{1}$ convolutional layer applied to the output of the last MFF to predict the final saliency map $S^f$. The total loss can be formulated mathematically as:
    \begin{equation}
        L_{total} = L_r(S^f, G) + \lambda(L_g(S^r, G) + L_g(S^d, G))
    \end{equation}
    where $ G $ is the ground truth and $\lambda$ is the balance factor between the global loss and the refined loss. We utilize the widely-used cross-entropy loss to calculate $ L_f $ and $ L_r $:
    \begin{equation}
        L(S, G) = -\sum_i[{G_ilog(S_i) + (1-G_i)log(1-S_i)}]
    \end{equation}
    where $i$ stands for the pixel index and $S$ is the prediction saliency map.
    
    \subsection{Implementation Details}
    Due to the characteristic of parameters sharing of the Siamese network, we replicate the depth maps to three channels as same as the RGB images. To be convenient for convolution, all RGB images and depth maps are resized to a fixed size of $320\times{320}$, the output size of the first VGG layer is also $320\times{320}$ for size recovering, and the subsequent output sizes of the DenseNet are $160\times{160}$, $80\times{80}$, $40\times{40}$, $20\times{20}$, and $20\times{20}$. Then, the hierarchical features from the encoder network are sent to the decoder network in a side-output way. 
    However, the resolutions and channel numbers of these features are varying, we design a Feature Channel Compression (FCC) module to unify the channels of the side-output features to a fixed number k. It brings two benefits: 1) reducing the cost of memory and computation, and 2) facilitating the element-wise operations of features between different levels.
    
    Our DMP modules are embedded in the high-level layer of the encoder network to propagate the context-aware message. We take the first DMP module as an example to explain the procedure of dynamic message propagation. 
    Firstly, we extract the RGB features from the $Conv4$ layer and hierarchical depth features from the $Conv3$, $Conv4$ and $Conv5$ layers. Depth features are aligned to the same size with the corresponding image features via the stride MaxPooling or interpolation operation. 
    Then, we transform the RGB features and multi-level depth features to a fixed channel number $C$ with a $1\times{1}$ convolution. 
    DMP takes RGB features as $F\in{R^{C\times{H\times{W}}}}$, we denote the initial feature map $H^{(0)}=F$. 
    The dynamic walk $\Delta{d}$ of image nodes is generated by a $3\times{3}$ convolutional layer. 
    Similarly, other $3\times{3}$ convolutional layers are applied on the depth feature maps to generate various dynamic walk $\Delta{\tilde{d}}$. 
    Affinity matrices and weights are calculated on the multi-level depth features $F^l_{dep}\in{R^{C\times{H\times{W}}}}$ via the $3\times{3}$ convolutional layer, where $l$ represents the $Conv3$, $Conv4$ and $Conv5$ layers. Then, we acquire the affinity matrices $A^l\in{R^{H\times{W\times{K\times{G}}}}}$ and filter weights $W^l\in{R^{H\times{W\times{K}}}}$, where $K$ is the sampling range (i.e., $3\times{3}$) and $K$ is the group size. The message $M\in{R^{C\times{H\times{W}}}}$ is calculated according to Eq.~\ref{eq5} and concatenated with $F$ to produce a refined feature map $H^{(1)}$, we only iterate once in this work to balance the performance and efficiency.

    \begin{table*}
    \renewcommand\arraystretch{1.5}
        \centering
        \fontsize{9}{12}\selectfont
        \caption{Comparison results on six challenging datasets with S-measure ($\mathcal{S}$), max F-measure ($\mathcal{F}$), max E-measure ($\mathcal{E}$) and MAE scores ($\mathcal{M}$). The symbol $\uparrow (\downarrow)$ means the score more higher (lower), more better. The best detection results are highlighted in boldface.}
        \label{tab:tab1}
        \large
        \resizebox{\textwidth}{!}{
        \begin{tabular}{c|c|cccc|cccc|cccc|cccc|cccc|cccc}
        \toprule[1.5pt]
        \multirow{2}{*}{Methods}& \multirow{2}{*}{Pub}&
        \multicolumn{4}{c|}{DUT} & \multicolumn{4}{c|}{NJU2K} & \multicolumn{4}{c|}{NLPR} & \multicolumn{4}{c|}{RGBD135} & \multicolumn{4}{c|}{SIP} & \multicolumn{4}{c}{STERE}\cr\cline{3-26}
        &  & $\mathcal{M}\downarrow$ & $\mathcal{F}\uparrow$ & $\mathcal{E}\uparrow$ & $\mathcal{S}\uparrow$ & $\mathcal{M}\downarrow$ & $\mathcal{F}\uparrow$ & $\mathcal{E}\uparrow$ & $\mathcal{S}\uparrow$ & $\mathcal{M}\downarrow$ & $\mathcal{F}\uparrow$ & $\mathcal{E}\uparrow$ & $\mathcal{S}\uparrow$ & $\mathcal{M}\downarrow$ & $\mathcal{F}\uparrow$ & $\mathcal{E}\uparrow$ & $\mathcal{S}\uparrow$ & $\mathcal{M}\downarrow$ & $\mathcal{F}\uparrow$ & $\mathcal{E}\uparrow$ & $\mathcal{S}\uparrow$ & $\mathcal{M}\downarrow$ & $\mathcal{F}\uparrow$ & $\mathcal{E}\uparrow$ & $\mathcal{S}\uparrow$ \cr
        \midrule[1pt]
        \multirow{4}{*}{}
                  \textbf{S2MA}\cite{9156287} & CVPR'20 & 0.043 & 0.886 & 0.921 & 0.903 & 0.053 & 0.889 & 0.930 & 0.894 & 0.030 & 0.847 & 0.937 & 0.915 & 0.021 & 0.893 & {\color{blue} \textbf{\underline{0.971}}} & {\color{red} \textbf{\underline{0.941}}} & 0.054 & 0.884 & 0.920 & 0.878 & 0.053 & 0.855 & 0.907 & 0.890 \\
                  \hline
                  \textbf{UCNet}\cite{9156838} & CVPR'20 & 0.086 & 0.773 & 0.855 & 0.797 & 0.043 & 0.886 & 0.930 & 0.897 & 0.025 & 0.886 & 0.951 & 0.920 & {\color{red} \textbf{\underline{0.019}}} & 0.905 & 0.967 & 0.934 & 0.051 & 0.879 & 0.919 & 0.875 & 0.039 & 0.885 & 0.922 & 0.903 \\
                  \hline
                  \textbf{JLDCF}\cite{9157592} & CVPR'20 & 0.043 & 0.882 & 0.931 & 0.906 & 0.043 & 0.903 & 0.944 & 0.903 & 0.022 & 0.916 & 0.962 & 0.925 & 0.022 & 0.919 & 0.968 & 0.929 & 0.051 & 0.885 & 0.923 & 0.879 & 0.042 & 0.901 & 0.946 & 0.905 \\
                  \hline
                  \textbf{SSF}\cite{9156645} & CVPR'20 & 0.034 & 0.915 & - & 0.915 & 0.043 & 0.896 & 0.935 & 0.899 & 0.026 & 0.896 & 0.953 & 0.914 & 0.025 & 0.883 & 0.941 & 0.905 & 0.053 & 0.880 & 0.921 & 0.874 & 0.044 & 0.890 & 0.936 & 0.893 \\
                  \hline
                  \textbf{D3Net}\cite{9107477} & TNNLS'20 & 0.095 & 0.740 & 0.833 & 0.775 & 0.046 & 0.900 & 0.939 & 0.900 & 0.030 & 0.897 & 0.953 & 0.912 & 0.031 & 0.885 & 0.946 & 0.898 & 0.063 & 0.861 & 0.909 & 0.860 & 0.046 & 0.891 & 0.938 & 0.899 \\
                  \hline
                  \textbf{CoNet}\cite{ji2020accurate} & ECCV'20 & 0.034 & 0.908 & 0.941 & 0.918 & 0.047 & 0.872 & 0.912 & 0.895 & 0.031 & 0.848 & 0.934 & 0.908 & 0.027 & 0.862 & 0.945 & 0.911 & 0.063 & 0.867 & 0.913 & 0.858 & 0.040 & 0.885 & 0.924 & 0.908 \\
                  \hline
                  \textbf{ATSA}\cite{asta} & ECCV'20 & 0.032 & 0.920 & 0.948 & 0.918 & 0.040 & 0.893 & 0.921 & 0.901 & 0.028 & 0.876 & 0.945 & 0.907 & 0.024 & 0.885 & 0.952 & 0.907 & 0.058 & 0.873 & 0.911 & 0.864 & 0.039 & 0.884 & 0.921 & 0.897 \\
                  \hline
                  \textbf{CMMS}\cite{li2020rgbd} & ECCV'20 & 0.037 & 0.906 & 0.940 & 0.913 & 0.044 & 0.886 & 0.914 & 0.900 & 0.027 & 0.869 & 0.945 & 0.915 & 0.020 & 0.922 & 0.970 & 0.932 & 0.061 & 0.871 & 0.910 & 0.867 & 0.043 & 0.879 & 0.922 & 0.895 \\
                  \hline
                  \textbf{BBSN}\cite{9562295} & ECCV'20 & 0.035 & 0.924 & 0.953 & 0.920 & 0.037 & 0.899 & 0.918 & 0.917 & 0.025 & 0.880 & 0.954 & 0.924 & 0.025 & 0.871 & 0.951 & 0.918 & 0.055 & 0.883 & 0.922 & 0.879 & 0.043 & 0.876 & 0.920 & 0.901 \\
                  \hline
                  \textbf{PGAR}\cite{chen2020progressively} & ECCV'20 & 0.035 & 0.913 & 0.944 & 0.919 & 0.042 & 0.893 & 0.916 & 0.909 & 0.025 & 0.883 & 0.954 & {\color{blue} \textbf{\underline{0.930}}} & 0.025 & 0.869 & 0.940 & 0.916 & 0.059 & 0.877 & 0.914 & 0.875 & 0.041 & 0.880 & 0.919 & {\color{red} \textbf{\underline{0.913}}} \\
                  \hline
                  \textbf{CMWN}\cite{li2020crossmodal} & ECCV'20 & 0.056 & 0.865 & 0.916 & 0.887 & 0.046 & 0.879 & 0.911 & 0.903 & 0.029 & 0.857 & 0.939 & 0.917 & 0.021 & 0.889 & 0.967 & 0.937 & 0.062 & 0.874 & 0.913 & 0.867 & 0.043 & 0.901 & 0.944 & 0.905 \\
                  \hline
                  \textbf{DANet}\cite{zhao2020single} & ECCV'20 & 0.043 & 0.883 & 0.934 & 0.899 & 0.045 & 0.871 & 0.922 & 0.899 & 0.028 & 0.870 & 0.949 & 0.915 & 0.023 & 0.887 & 0.967 & 0.924 & 0.054 & 0.876 & 0.918 & 0.875 & 0.043 & 0.892 & 0.937 & 0.901 \\
                  \hline
                  \textbf{HDFN}\cite{pang2020hierarchical} & ECCV'20 & 0.040 & 0.865 & 0.938 & 0.905 & 0.051 & 0.847 & 0.920 & 0.885 & 0.031 & 0.839 & 0.942 & 0.898 & 0.030 & 0.843 & 0.944 & 0.899 & 0.050 & 0.835 & 0.920 & 0.878 & 0.039 & 0.863 & 0.937 & 0.906 \\
                  \hline
                  \textbf{DSAF}\cite{sun2021deep} & CVPR'21 & 0.030 & 0.926 & 0.950 & 0.921 & 0.039 & 0.901 & 0.923 & 0.903 & 0.024 & 0.897 & 0.950 & 0.918 & 0.021 & 0.896 & 0.962 & 0.920 & - & - & - & - & {\color{blue} \textbf{\underline{0.036}}} & 0.898 & 0.933 & 0.904 \\
                  \hline
                  \textbf{UTA}\cite{9529069} & TIP'21 & - & - & - & - & 0.037 & 0.906 & 0.946 & 0.902 & {\color{blue} \textbf{\underline{0.020}}} & {\color{blue} \textbf{\underline{0.926}}} & {\color{blue} \textbf{\underline{0.965}}} & 0.928 & 0.026 & 0.897 & 0.933 & 0.901 & {\color{blue} \textbf{\underline{0.048}}} & 0.884 & {\color{blue} \textbf{\underline{0.926}}} & 0.873 & {\color{red} \textbf{\underline{0.033}}} & {\color{red} \textbf{\underline{0.912}}} & {\color{blue} \textbf{\underline{0.949}}} & {\color{blue} \textbf{\underline{0.910}}}\\
                  \hline
                  \textbf{CDINet}\cite{zhang2021crossmodality} & ACM'21 & {\color{blue} \textbf{\underline{0.029}}} & {\color{blue} \textbf{\underline{0.935}}} & {\color{blue} \textbf{\underline{0.957}}} & {\color{blue} \textbf{\underline{0.927}}} & {\color{blue} \textbf{\underline{0.036}}} & {\color{blue} \textbf{\underline{0.921}}} & {\color{blue} \textbf{\underline{0.951}}} & {\color{blue} \textbf{\underline{0.918}}} & 0.024 & 0.916 & 0.960 & 0.927 & 0.020 & {\color{red} \textbf{\underline{0.934}}} & 0.970 & {\color{blue} \textbf{\underline{0.937}}} & 0.054 & 0.884 &  0.915 & 0.875 & - & - & - & - \\
                  \hline
                  \textbf{DFMNet}\cite{zhang2021depth} & ACM'21 & - & - & - & - & 0.042 & 0.910 & 0.947 & 0.906 & 0.026 & 0.908 & 0.957 & 0.923 & 0.021 & 0.922 & {\color{red} \textbf{\underline{0.972}}} & 0.931 & 0.051 & {\color{blue} \textbf{\underline{0.887}}} & {\color{blue} \textbf{\underline{0.926}}} & {\color{blue} \textbf{\underline{0.883}}} & 0.045 & 0.893 & 0.941 & 0.898
                  \cr\midrule[1pt]
                  \textbf{Ours} & - & {\color{red} \textbf{\underline{0.027}}} & {\color{red} \textbf{\underline{0.938}}} & {\color{red} \textbf{\underline{0.959}}} & {\color{red} \textbf{\underline{0.933}}} & {\color{red} \textbf{\underline{0.035}}} & {\color{red} \textbf{\underline{0.922}}} & {\color{red} \textbf{\underline{0.954}}} & {\color{red} \textbf{\underline{0.921}}} & {\color{red} \textbf{\underline{0.019}}} & {\color{red} \textbf{\underline{0.929}}} & {\color{red} \textbf{\underline{0.969}}} & {\color{red} \textbf{\underline{0.937}}} & {\color{red} \textbf{\underline{0.019}}} & {\color{blue} \textbf{\underline{0.929}}}& 0.969 & {\color{blue} \textbf{\underline{0.937}}} & {\color{red} \textbf{\underline{0.044}}} & {\color{red} \textbf{\underline{0.908}}} & {\color{red} \textbf{\underline{0.936}}} & {\color{red} \textbf{\underline{0.895}}} & 0.037 & {\color{blue} \textbf{\underline{0.910}}} & {\color{red} \textbf{\underline{0.950}}} & {\color{red} \textbf{\underline{0.913}}} \cr
        \bottomrule[1.5pt]
        \end{tabular}
        }
    \end{table*}
     
    \section{Experiments}
    \label{sec:exp}
    \subsection{Datasets and Metrics}
    We carry out our experiments on the six challenging datasets: NJU2K~\cite{7025222}, NLPR~\cite{nlpr}, DUT~\cite{9010728}, RGBD135~\cite{rgbd135}, SIP~\cite{9107477} and STERE~\cite{6247708}. \textbf{NJU2K} contains 1985 RGB-D image pairs with diverse and complicated scenes, where depth maps are estimated from the stereo images. \textbf{NLPR} consists of 1000 RGB-D images captured by the Kinect, most of them have more than one salient object. \textbf{DUT} includes 1200 paired RGB-D images from more challenging scene such as transparent objects and low-intensity environments. \textbf{RGBD135} is a small dataset, which only contains 135 paired indoor images. \textbf{SIP} consists of 929 person RGBD image pairs captured by Huawei Meta10 with high-resolution. \textbf{STERE} includes 1000 image pairs downloaded from the Internet, which is the first stereoscopic image collection.
    
    We divide these datasets into the training data and the testing data. Specifically, the training set contains 1500 samples from NJU2K, 700 samples from NLPR and 800 samples from DUT. The testing set includes the rest of NJU2K (485), NLPR (300) and DUT (400), and the whole of RGBD135 (135), SIP (929) and STERE (1000). 
    
    We apply the widely used metrics including F-measure ($\mathcal{F}$)~\cite{5206596}, mean absolute error ($\mathcal{M}$)~\cite{7293665}, S-measure ($\mathcal{S}$)~\cite{8237749} and E-measure ($\mathcal{E}$)~\cite{fan2018enhancedalignment} to evaluate the performance of our method and the state-of-the-art methods. The $\mathcal{F}$ considers the balance between precision and recall, which can be calculated by comparing the saliency map with the ground truth. This can be formulated as:
    \begin{equation}
        \mathcal{F} = \frac{(1 + \beta^2)\cdot{Precision\cdot{Recall}}}{\beta^2\cdot{Precision + Recall}}
    \end{equation}
    where $\beta^2$ is usually set to 0.3 to emphasize the precision. 
    
    The $\mathcal{M}$ computes the average absolute difference of per-pixel between the saliency map and the ground truth:
    \begin{equation}
        \mathcal{M} = \frac{1}{H\times{W}}\sum^H_{y=1}\sum^W_{x=1}{|S(x,y)-G(x,y)|}
    \end{equation}
    where $S$ is the saliency map, $G$ is the ground truth, $H$ and $W$ are the height and width of sailency map respectively. 
    
    The $\mathcal{S}$ is proposed to evalute the similarity of the structure between the saliency map and the ground truth, which is defined as:
    \begin{equation}
        \mathcal{S} =\alpha * S_o + (1 - \alpha) * S_r
    \end{equation}
    where $\alpha$ is set to 0.5 balance the object-wise structural similarity ($S_o$) and the region-wise structural similarity ($S_r$).
    
    The $\mathcal{E}$ simultaneously captures global statistics and local pixel information to evaluate the saliency map:
    \begin{equation}
        \mathcal{E} = \frac{1}{W\times{H}}\sum^H_{y=1}\sum^W_{x=1}{\Phi_{FM}{(x,y)}}
    \end{equation}
    where $H$ and $W$ are the height and width of sailency map respectively, and $\Phi_{FM}$ is the enhanced alignment matrix.
    
    \begin{figure*}[t]
        \centering
        \includegraphics[width=\linewidth]{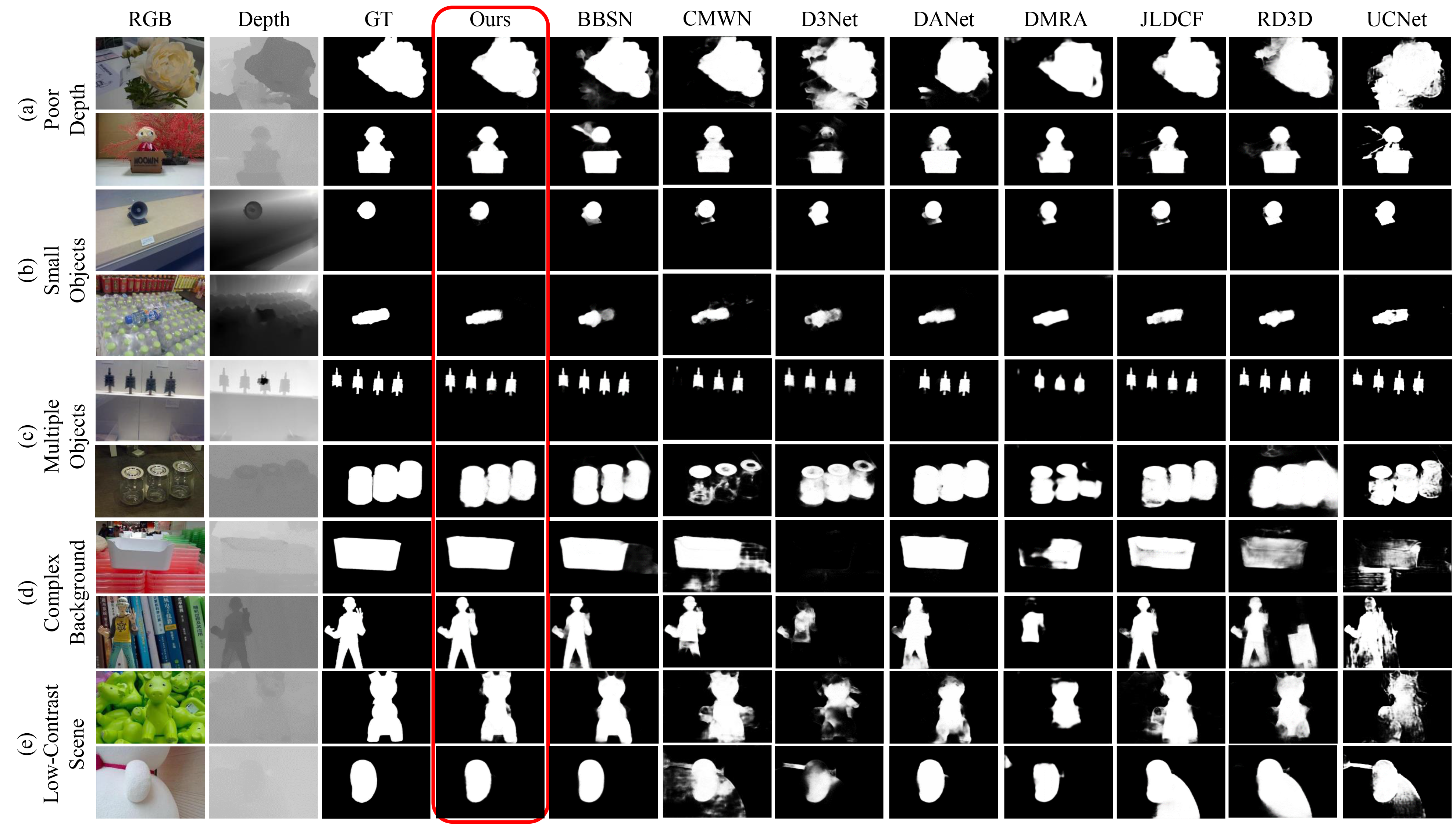}
        \caption{Qualitative visual comparison of our model with eight SOTA models under various scenes.}
        \label{fig:fig5}
    \end{figure*}
    
    \subsection{Training Details}
    We implement our DMP-Net with PyTorch on a GeForce RTX 2070 GPU. Parameters of the backbone network (the first convolutional layer of VGG-16 \cite{simonyan2014very} and DenseNet-161 \cite{huang2017densely}) are initialized from the models pre-trained on ImageNet. We extract the side-out features of six convolutional layers as multi-level features. The RGB and depth streams share the same weights and have the same architecture. The unified number $ k $ of channels in FCC is set to $ 64 $. Other parameters are initialized using the default PyTorch settings. $\lambda$ in the total loss function is set to $16 \times 16$ to balance the number variance of pixels between the global prediction map and the refined prediction map. The Adam algorithm is used to optimize our model. We set the initial learning rate to 5e-5 and weight decay to 5e-4. The input RGB images and depth maps are resized to $ 320 \times 320 $. We augment all the training images by horizontal flipping and random cropping. It takes about thirty hours to train the model with a batch size of 1 for 45 epochs.
    
    \subsection{Backbone Details}
    We use four different backbones to extract features in the encoder network, which are ResNet101, ResNet50, DenseNet, and VGG16, respectively. Before feeding to the backbone, we fix the input size of the RGB-D images as $320\times{320}$.
    
    \textbf{VGG16:} We remove the fully-connected layers of VGG16, the rest of it also has 13 convolutional layers and 5 max-pooling layers, which are divided into 6 blocks: $conv1\_2$, $conv2\_2$, $conv3\_3$, $conv4\_3$, $conv5\_3$, and $pool\_5$. Each block has a $side~path$ with extra convolutional layers to connect with the corresponding block in the decoder network. To preserve the resolution of the coarse features from the \\$side~path6$, we change the stride of $pool5$ to 1 and use a dilated convolution with a rate of 2 instead for the side convolutional layers. The coarsest features generated from the modified VGG16 have a size of $20\times{20}$.
    
    \textbf{ResNet:} The spatial size of coarsest features is also $20\times{20}$, which is the same as VGG16. Since the first convolutional layer of ResNet has a stride of 2, the output features have a spatial size of $160\times{160}$. However, we need to obtain the full size ($320\times{320}$) features for size recovering in the decoder, so that we employ the $conv1$ layers of VGG16 as the first block for feature extraction. The $conv1$, $conv2\_3$, $conv3\_4$, $conv4\_23$, and $conv5\_3$ of ResNet101 are connected to the $side ~path1\sim6$. The only difference between ResNet50 and ResNet101 is that the forth convolutional layer of ResNet50 only has 6 residual blocks.
    
    \textbf{DenseNet:} The structure of DenseNet is similar to ResNet. The slight difference is the fundamental components that are changed from the residual block to the dense block. Therefore, we adopt the first layer of VGG16 to extract the full size features. Subsequently, the $conv1$, $dense1\_6$, $dense2\_12$, $dense3\_36$, and $dense4\_24$ of DenseNet161 are connected to the $side ~path1\sim6$.
    
    \subsection{Comparison with SOTAs}
    We compare our method with seventeen state-of-the-art methods, including TANet~\cite{8603756}, CPFP~\cite{8953818}, DMRA~\cite{9010728}, S2MA~\cite{9156287}, UCNet~\cite{9156838}, JLDCF~\cite{9157592}, SSF~\cite{9156645}, D3Net~\cite{9107477}, CoNet~\cite{ji2020accurate}, ATSA~\cite{asta}, CMMS~\cite{li2020rgbd}, BBSNet~\cite{9562295}, PGAR~\cite{chen2020progressively}, CMWNet~\cite{li2020crossmodal}, DANet~\cite{zhao2020single}, HDFNet~\cite{pang2020hierarchical} and DSAF~\cite{sun2021deep}. The above experiment results are from the published papers and the released codes which we re-train and re-test with their default settings.
    
    \begin{table*}
    \renewcommand\arraystretch{1.5}
        \centering
        \fontsize{9}{12}\selectfont
        \caption{Quantitative evaluation for the first step ablation study that only replaces DenseNet with other backbones.}
        \label{tab:tab2}
        \large
        \resizebox{\textwidth}{!}{
        \begin{tabular}{c|cccc|cccc|cccc|cccc|cccc|cccc}
        \toprule[1.5pt]
        \multirow{2}{*}{Methods}&
        \multicolumn{4}{c|}{DUT} & \multicolumn{4}{c|}{NJU2K} & \multicolumn{4}{c|}{NLPR} & \multicolumn{4}{c|}{RGBD135} & \multicolumn{4}{c|}{SIP} & \multicolumn{4}{c}{STERE}\cr\cline{2-25}
          & $\mathcal{M}\downarrow$ & $\mathcal{F}\uparrow$ & $\mathcal{E}\uparrow$ & $\mathcal{S}\uparrow$ & $\mathcal{M}\downarrow$ & $\mathcal{F}\uparrow$ & $\mathcal{E}\uparrow$ & $\mathcal{S}\uparrow$ & $\mathcal{M}\downarrow$ & $\mathcal{F}\uparrow$ & $\mathcal{E}\uparrow$ & $\mathcal{S}\uparrow$ & $\mathcal{M}\downarrow$ & $\mathcal{F}\uparrow$ & $\mathcal{E}\uparrow$ & $\mathcal{S}\uparrow$ & $\mathcal{M}\downarrow$ & $\mathcal{F}\uparrow$ & $\mathcal{E}\uparrow$ & $\mathcal{S}\uparrow$ & $\mathcal{M}\downarrow$ & $\mathcal{F}\uparrow$ & $\mathcal{E}\uparrow$ & $\mathcal{S}\uparrow$ \cr
        \midrule[1pt]
        \multirow{4}{*}{}
                  \textbf{DMP-Net (VGG-16)} & 0.031 & 0.927 & 0.953 & 0.923 & 0.039 & 0.913 & 0.946 & 0.910 & 0.023 & 0.917 & 0.963 & 0.927 & 0.022 & 0.911 & 0.950 & 0.919 & 0.048 & 0.894 & 0.928 & 0.884 & 0.041 & 0.902 & 0.943 & 0.903 \cr\cline{1-25}
                  \textbf{DMP-Net (ResNet-50)} & 0.036 & 0.915 & 0.946 & 0.915 & 0.041 & 0.916 & 0.946 & 0.910 & 0.026 & 0.902 & 0.954 & 0.920 & 0.025 & 0.907 & 0.947 & 0.921 & 0.051 & 0.894 & 0.925 & 0.882 & 0.049 & 0.885 & 0.934 & 0.892 \cr\cline{1-25}
                  \textbf{DMP-Net (ResNet-101)} & 0.030 & 0.935 & \textbf{0.959} & 0.930 & 0.039 & 0.915 & 0.949 & 0.915 & 0.022 & 0.922 & 0.966 & 0.933 & \textbf{0.019} & \textbf{0.932} & \textbf{0.974} & \textbf{0.939} & 0.050 & 0.895 & 0.928 & 0.885 & 0.043 & 0.902 & 0.945 & 0.907 \cr\cline{1-25}
                  \textbf{DMP-Net (DenseNet-161)} & \textbf{0.027} & \textbf{0.938} & \textbf{0.959} & \textbf{0.933} & \textbf{0.035} & \textbf{0.922} & \textbf{0.954} & \textbf{0.921} & \textbf{0.019} & \textbf{0.929} & \textbf{0.969} & \textbf{0.937} & \textbf{0.019} & 0.929 & 0.969 & 0.937 & \textbf{0.044} & \textbf{0.908} & \textbf{0.936} & \textbf{0.895} & \textbf{0.037} & \textbf{0.910} & \textbf{0.950} & \textbf{0.913} \cr
        \bottomrule[1.5pt]
        \end{tabular}
        }
    \end{table*}
    
    \begin{figure*}[t]
        \centering
        \includegraphics[width=\linewidth]{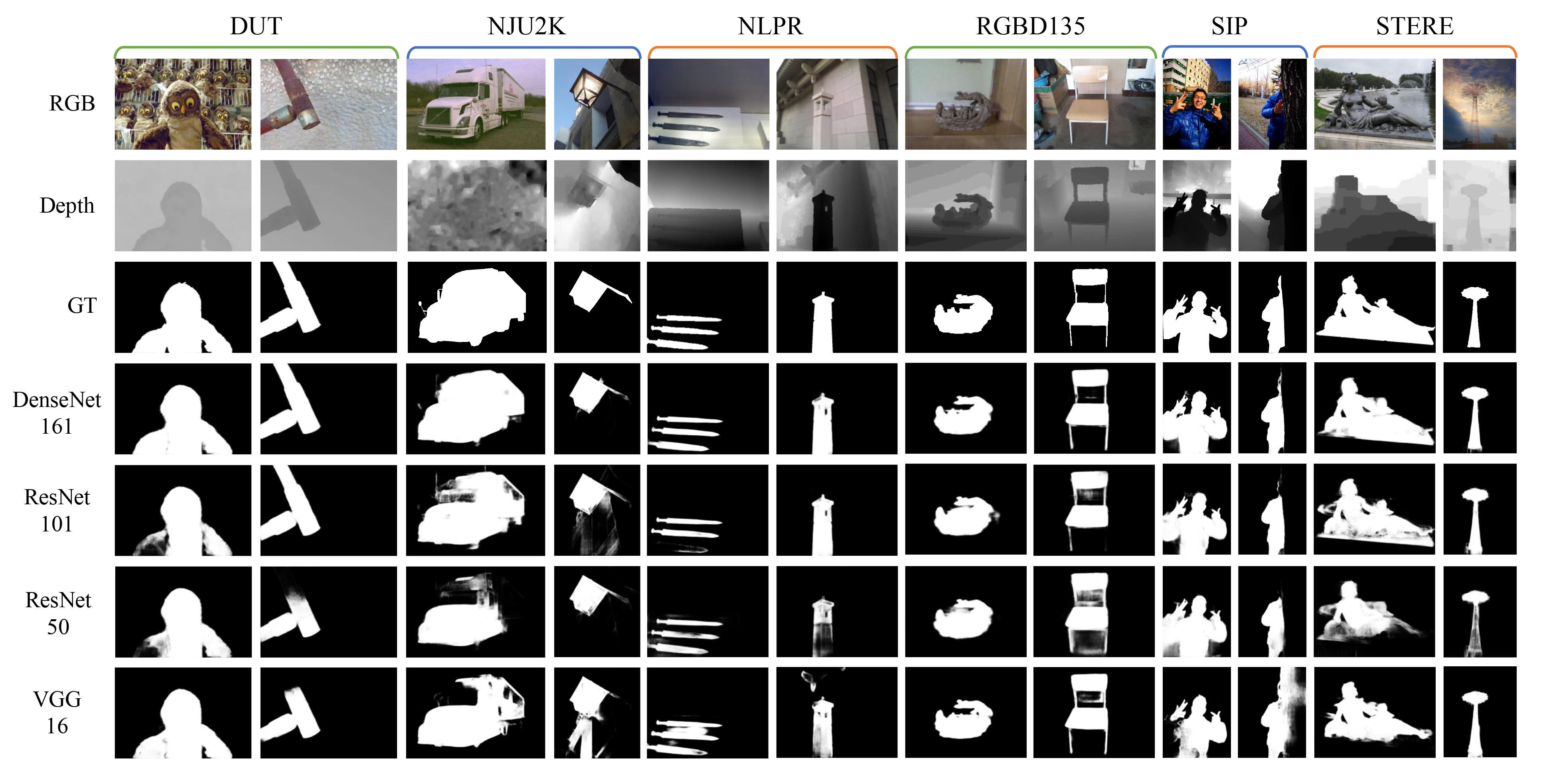}
        \caption{Visual examples from DUT, NJU2K, NLPR, RGBD135, SIP and STERE datasets for ablation studies of different backbones.}
        \label{fig:fig7}
    \end{figure*}
    
    \textbf{Quantitative Results.} As shown in Tab.~\ref{tab:tab1}, our DMP-Net consistently outperforms SOTAs over four evaluation metrics: MAE ($\mathcal{M}$), F-measure ($\mathcal{F}$), E-measure ($\mathcal{E}$), and S-measure ($\mathcal{S}$). 
    The outstanding performance of our DMP-Net over SOTAs is remarkable, demonstrating that DMP-Net can well excavate the beneficial features from RGB images and depth maps, and integrate multi-modal and multi-level features.
    We achieve almost all best or sub-optimal results on the six challenging datasets under four metrics. Especially for DUT and SIP datasets, performance gains over the sub-optimal method are (0.2\% $\sim$ 0.4\%, 0.3\% $\sim$ 1.2\%, 0.2\% $\sim$ 1\%, 0.6\% $\sim$ 1.2\%) for the metrics ($\mathcal{M}$, $\mathcal{F}$, $\mathcal{E}$ and $\mathcal{S}$).
    
    \begin{figure*}[t]
        \centering
        \includegraphics[width=\linewidth]{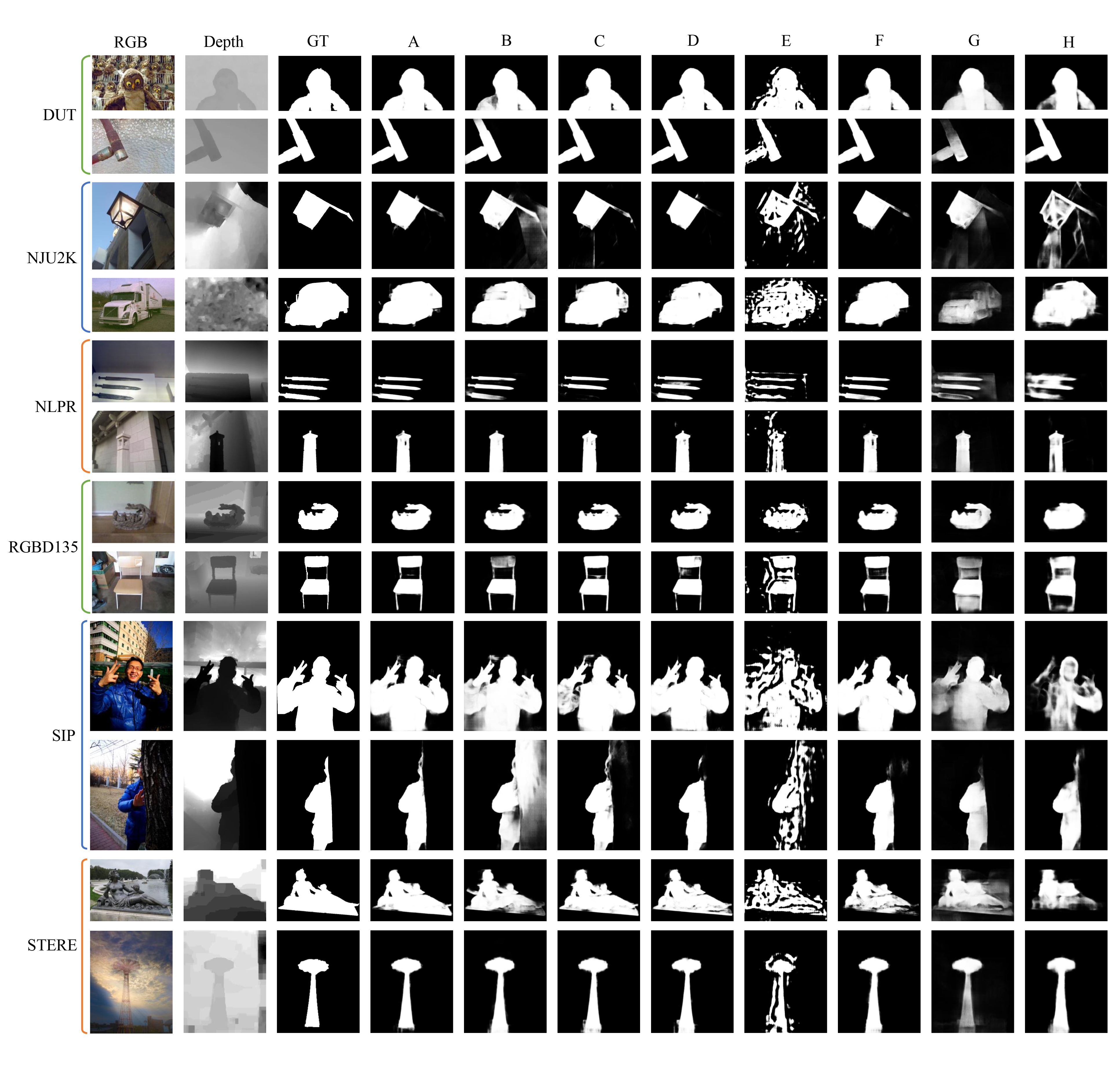}
        \caption{Visual examples from DUT, NJU2K, NLPR, RGBD135, SIP and STERE datasets for ablation studies. For different configurations, “A”: Ours (\textit{DenseNet161+full modules}), “B”: Ours (\textit{DenseNet161+w/o DMP}), “C”: Ours (\textit{DenseNet161+w/ one DMP}), “D”: Ours (\textit{DenseNet161+w/ three DMP}), “E”: Ours (\textit{DenseNet161+w/o Inception}), “F”: Ours (\textit{DenseNet161+w/o Dense link}), “G”: Ours (\textit{DenseNet161+w/ only RGB}), “H”: Ours (\textit{DenseNet161+w/ only D}).}
        \label{fig:fig7}
    \end{figure*}
    
    \begin{table*}[h]
    \renewcommand\arraystretch{1.5}
        \centering
        \caption{\label{tab:tab3}
        Quantitative evaluation for the second step ablation study that removes the different components.}
        \fontsize{9}{12}\selectfont
        \large
        \resizebox{\textwidth}{!}{
        \begin{tabular}{c|cccc|cccc|cccc|cccc|cccc|cccc}
        \toprule[1.5pt]
        \multirow{2}{*}{Methods}&
        \multicolumn{4}{c|}{DUT} & \multicolumn{4}{c|}{NJU2K} & \multicolumn{4}{c|}{NLPR} & \multicolumn{4}{c|}{RGBD135} & \multicolumn{4}{c|}{SIP} & \multicolumn{4}{c}{STERE} \\
        \cline{2-25}
          & $\mathcal{M}\downarrow$ & $\mathcal{F}\uparrow$ & $\mathcal{E}\uparrow$ & $\mathcal{S}\uparrow$ & $\mathcal{M}\downarrow$ & $\mathcal{F}\uparrow$ & $\mathcal{E}\uparrow$ & $\mathcal{S}\uparrow$ & $\mathcal{M}\downarrow$ & $\mathcal{F}\uparrow$ & $\mathcal{E}\uparrow$ & $\mathcal{S}\uparrow$ & $\mathcal{M}\downarrow$ & $\mathcal{F}\uparrow$ & $\mathcal{E}\uparrow$ & $\mathcal{S}\uparrow$ & $\mathcal{M}\downarrow$ & $\mathcal{F}\uparrow$ & $\mathcal{E}\uparrow$ & $\mathcal{S}\uparrow$ & $\mathcal{M}\downarrow$ & $\mathcal{F}\uparrow$ & $\mathcal{E}\uparrow$ & $\mathcal{S}\uparrow$ \\
          \hline
                  \textbf{DMP-Net (w/o DMP)} & 0.034 & 0.929 & 0.953 & 0.924 & 0.044 & 0.911 & 0.944 & 0.910 & 0.025 & 0.918 & 0.964 & 0.927 & 0.023 & 0.917 & 0.954 & 0.929 & 0.051 & 0.896 & 0.932 & 0.886 & 0.044 & 0.900 & 0.944 & 0.907 \\
                  \hline
                  \textbf{DMP-Net (w/ one DMP)} & 0.028 & 0.938 & 0.959 & 0.932 & 0.036 & 0.920 & 0.953 & 0.918 & 0.020 & 0.926 & 0.968 & 0.934 & 0.020 & 0.923 & 0.960 & 0.932 & 0.045 & 0.904 & 0.934 & 0.892 & 0.039 & 0.908 & 0.949 & 0.911 \\
                  \hline
                  \textbf{DMP-Net (w/ three DMP)} & \textbf{0.027} & 0.940 & \textbf{0.962} & \textbf{0.936} & 0.036 & 0.921 & 0.953 & 0.918 & 0.020 & 0.924 & \textbf{0.969} & 0.934 & \textbf{0.019} & \textbf{0.932} & 0.968 & 0.938 & 0.046 & 0.898 & 0.931 & 0.889 & 0.039 & 0.906 & 0.948 & 0.910 \\
                  \hline
                  \textbf{DMP-Net (w/o Inception)} & 0.066 & 0.841 & 0.918 & 0.841 & 0.098 & 0.796 & 0.892 & 0.785 & 0.047 & 0.825 & 0.939 & 0.848 & 0.042 & 0.835 & 0.952 & 0.865 & 0.105 & 0.748 & 0.876 & 0.763 & 0.100 & 0.760 & 0.888 & 0.774 \\
                  \hline
                  \textbf{DMP-Net (w/o Dense link)} & \textbf{0.027} & \textbf{0.941} & 0.961 & 0.934 & 0.038 & 0.918 & 0.951 & 0.916 & 0.021 & 0.928 & 0.967 & 0.935 & \textbf{0.019} & 0.930 & 0.967 & \textbf{0.939} & 0.045 & 0.900 & 0.933 & 0.892 & 0.039 & 0.907 & 0.948 & 0.911 \\
                  \hline
                  \textbf{DMP-Net (w/ only RGB)} & 0.093 & 0.864 & 0.916 & 0.838 & 0.085 & 0.869 & 0.912 & 0.866 & 0.049 & 0.872 & 0.933 & 0.887 & 0.039 & 0.886 & 0.941 & 0.903 & 0.085 & 0.854 & 0.905 & 0.853 & 0.099 & 0.834 & 0.900 & 0.845 \\
                  \hline
                  \textbf{DMP-Net (w/ only D)} & 0.050 & 0.883 & 0.931 & 0.891 & 0.063 & 0.873 & 0.923 & 0.877 & 0.035 & 0.880 & 0.947 & 0.901 & 0.028 & 0.900 & 0.959 & 0.916 & 0.064 & 0.870 & 0.915 & 0.865 & 0.078 & 0.818 & 0.901 & 0.839 \\
                  \hline
                  \textbf{DMP-Net (full)} & \textbf{0.027} & 0.938 & 0.959 & 0.933 & \textbf{0.035} & \textbf{0.922} & \textbf{0.954} & \textbf{0.921} & \textbf{0.019} & \textbf{0.929} & \textbf{0.969} & \textbf{0.937} & \textbf{0.019} & 0.929 & \textbf{0.969} & 0.937 & \textbf{0.044} & \textbf{0.908} & \textbf{0.936} & \textbf{0.895} & \textbf{0.037} & \textbf{0.910} & \textbf{0.950} & \textbf{0.913} \\
        \bottomrule[1.5pt]
        \end{tabular}
        }
    \end{table*}
    
    \textbf{Visual Comparison.} We select five challenging scenes from the datasets, including (a) poor-depth scene, (b) small-object scene, (c) multi-object scene, (d) complex-background scene, and (e) low-contrast scene. We visualize the results of our DMP-Net and eight SOTAs in Fig.~\ref{fig:fig5}:
    
    Poor quality depth maps in scene \textbf{(a)}. The SOTAs, such as D3Net and UCNet, cannot alleviate the interference from the poor depth maps, leading to undesirable prediction results. In contrast, our DMP-Net eliminates the misleading depth information and exactly detects the salient object by dynamical message propagation. 
    
    Small but salient objects in scene \textbf{(b)}. For example, the loudspeaker and water bottle are too small to completely detect by other methods, except our DMP-Net. 
    
    Multiple salient objects in  scene \textbf{(c)}. Our method can recognize all salient objects in the scene and generate more sharper edges than the SOTAs. It is worth noticed that even the depth map lacks clear foreground information in the second row of (c), our algorithm can still detect the salient objects correctly. 
    
    Complex backgrounds in scene \textbf{(d)}. The complex background of RGB images potentially disturbs the prediction of salient objects. We can clearly observe that the SOTAs either only detect parts of salient objects or wrongly recognize the background regions as salient objects. In contrast, our DMP-Net can overcome the complex background from RGB images, thus detecting salient objects with their boundaries well preserved.  
    
    Low contrast between the salient object and its background in scene \textbf{(e)}. Most methods fail to segment salient object out of the entire scene, but our method accurately predicts salient object by dynamically exploiting depth information to identify the background region.
    
    
    \subsection{Ablation Study}
    We conduct comprehensive ablation studies by replacing the backbone or removing the components of the proposed network. We set the DMP-Net with full components as a reference.
    
    \textbf{Efficiency of the backbone network.} We first replace DenseNet161 with VGG and different versions of ResNet to show its strong ability of feature extraction. As is shown in Tab.~\ref{tab:tab2}, DenseNet161 achieves the best performance in almost all testing datasets over four metrics. However, other backbones perform unsatisfactorily. The reason behind this phenomenon is that deep convolutional layers can excavate abundant semantic information and dense skip connection can enhance feature aggregation and alleviate vanishing gradient problem.
    
    \textbf{Efficiency of Dynamic Message Propagation.} In order to demonstrate the significance of DMP, which can dynamically aggregate cross-modal features, we carry out thorough experiments on the number of DMP module. For ease of notation, we denote the model without DMP module as (\textit{DPM-Net (w/o DMP)}), the model that only retains DMP module in the fifth encoder layer as (\textit{DMP-Net (w/ one DMP)}) and the model which obtains DMP module in the last three layers as (\textit{DMP-Net (w/ three DMP)}). From Tab.~\ref{tab:tab3}, it is obvious that the performance of DMP-Net is boosted dramatically when the number of DMP modules increases. However, when the number of DMP modules reaches three, the module has degraded results. The reason behind phenomenon that too many DMP modules repeatedly extract the middle-level depth information leading the information redundancy, and low-level features carry some noise that may interfere with sampling the context-aware nodes. It is worth noticed that DMP module has the advantage of efficiently making cross-modal fusion and generating hybrid semantic features.
    
    \textbf{Efficiency of Multi-modal Multi-scale Feature Fusion.} To evaluate the efficiency of MFF module, we first remove the inception-like part in MFF module and then delete dense links, which are denoted as (\textit{DMP-Net (w/o Inception)}) and (\textit{DMP-Net (w/o Dense link)}), respectively. Tab.~\ref{tab:tab3} reflects that without the inception-like part that extends the receptive filed, only integrating multi-level features with a simple element-wise addition will result in insufficient fusion of multi-level features and performance degradation. Additionally, when we replace dense links with single skip connections, the performance also declines to a certain extent. RGB features and Depth features have significant complementarity. RGB images include textual and semantic cues while depth maps contain geometric information, when one of them has poor quality, another can provide an effective supplement. Thereby the model with single input (\textit{DPM-Net (w/ only RGB)} and \textit{DPM-Net (w/ only D)}) has poor performance due to lack of the assistance of another modality.
    
    
    
    \begin{figure}
        \centering
        \includegraphics[width=\linewidth]{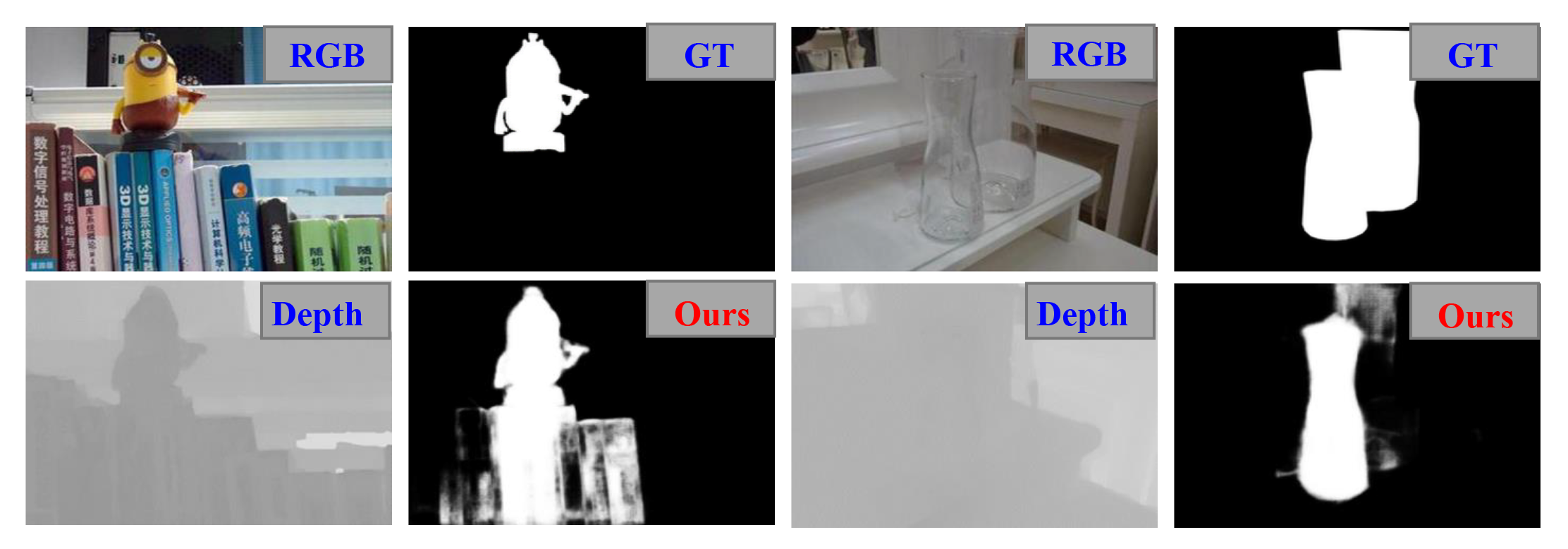}
        \caption{Deficiency examples of our DMP-Net.}
        \label{fig:fig6}
    \end{figure}
    
    \subsection{Deficiency}
    Although DMP-Net demonstrates superiority across testing datasets, there still exist some tricky scenes which most current SOTAs also cannot solve well. As illustrated in Fig.~\ref{fig:fig6}, the salient object in the left example is small while the books under it are conspicuous and cover a large part in the picture, and the depth map also tends to form the minion and books to an entirety. It is quite misleading that our method incorrectly predicts part of the books as a salient object. In the right example, the salient objects are transparent in the RGB image and the quality of the depth map is very poor, thus DMP-Net only detects the front glass bottle, ignoring the behind one. The dynamic message propagates algorithm samples the relevant nodes for message passing, some sampled nodes perhaps from the background if the salient objects and background objects have a similar depth. While the misleading depth nodes also interfere with the RGB features by message passing for salient object detection. It is concluded that if the RGB image and depth map are too confusing to extract useful information to detect salient objects, the performance of our method will be satisfactory.
        
    \section{Conclusion}
    \label{sec:con}
    In this paper, we propose a dynamic message propagation network, dubbed DMP-Net, which innovatively applies a graph-based paradigm to the long-range contextual dependencies learning for RGB-D SOD. DMP-Net consists of three key components. First, A Siamese encoder is utilized to extract multi-level features of RGB image and depth map simultaneously. Then, we propose a Dynamic Message Propagation (DMP) module to dynamically aggregate cross-modal features and enhance the hybrid information. Moreover, we design a Multi-level Feature Fusion (MFF) module to integrate the higher-level features and obtain the fine-grained features step by step. Extensive quantitative and qualitative experiments demonstrate the superiority of DMP-Net over SOTA methods and its robustness and generalization ability to challenging cases.
    
    \bibliographystyle{IEEEtran} 
    \bibliography{ref}

    
    %

    



    \ifCLASSOPTIONcaptionsoff
      \newpage
    \fi

\end{document}